\pdfoutput=1

\documentclass[11pt]{article}

\PassOptionsToPackage{dvipsnames}{xcolor}
\usepackage[final]{acl}
\usepackage{amsmath}
\usepackage{times}
\usepackage{latexsym}
\usepackage[T1]{fontenc}

\usepackage[utf8]{inputenc}

\usepackage{microtype}

\usepackage{inconsolata}
\usepackage[dvipsnames]{xcolor}

\usepackage{graphicx}
\usepackage{amssymb}
\usepackage{booktabs}
\usepackage{multirow}
\usepackage{multicol}

\title{ESPnet-SpeechLM: An Open Speech Language Model Toolkit}

\author{
 \textbf{Jinchuan Tian\textsuperscript{1}}\quad
 \textbf{Jiatong Shi\textsuperscript{1}}\quad
 \textbf{William Chen\textsuperscript{1}}\quad
 \textbf{Siddhant Arora\textsuperscript{1}}\quad
 \\
 \textbf{Yoshiki Masuyama\textsuperscript{2}}\quad
 \textbf{Takashi Maekaku\textsuperscript{3}}\quad
 \textbf{Yihan Wu\textsuperscript{1,4}}\quad
 \textbf{Junyi Peng\textsuperscript{1,5}}\quad
 \\
 \textbf{Shikhar Bharadwaj\textsuperscript{1}}\quad
 \textbf{Yiwen Zhao\textsuperscript{1}}\quad
 \textbf{Samuele Cornell\textsuperscript{1}}\quad
 \textbf{Yifan Peng\textsuperscript{1}}\quad
  \\
  \textbf{Xiang Yue\textsuperscript{1}}\quad
  \textbf{Chao-Han Huck Yang\textsuperscript{6}}\quad
  \textbf{Graham Neubig\textsuperscript{1}}\quad
  \textbf{Shinji Watanabe\textsuperscript{1}}
\\
 \textsuperscript{1}Carnegie Mellon University\quad
 \textsuperscript{2}Mitsubishi Electric Research Laboratories\quad
 \textsuperscript{3}LY Corporation\quad
 \\
 \textsuperscript{4}Renmin University of China\quad
 \textsuperscript{5}Brno University of Technology\quad \textsuperscript{6}NVIDIA Research
\\
 \small{
   \textbf{Correspondence:} \href{jinchuat@andrew.cmu.edu}{jinchuat@andrew.cmu.edu}
 }
}

\begin{document}
\maketitle
\begin{abstract}
We present ESPnet-SpeechLM, an open toolkit designed to democratize the development of speech language models (SpeechLMs) and voice-driven agentic applications. The toolkit standardizes speech processing tasks by framing them as universal sequential modeling problems, encompassing a cohesive workflow of data preprocessing, pre-training, inference, and task evaluation. With ESPnet-SpeechLM, users can easily define task templates and configure key settings, enabling seamless and streamlined SpeechLM development. The toolkit ensures flexibility, efficiency, and scalability by offering highly configurable modules for every stage of the workflow.
To illustrate its capabilities, we provide multiple use cases demonstrating how competitive SpeechLMs can be constructed with ESPnet-SpeechLM, including a 1.7B-parameter model pre-trained on both text and speech tasks, across diverse benchmarks. The toolkit and its recipes are fully transparent and reproducible at: \url{https://github.com/espnet/espnet/tree/speechlm}.

\end{abstract}

\section{Introduction}

\begin{figure}
    \centering
    \includegraphics[width=\linewidth]{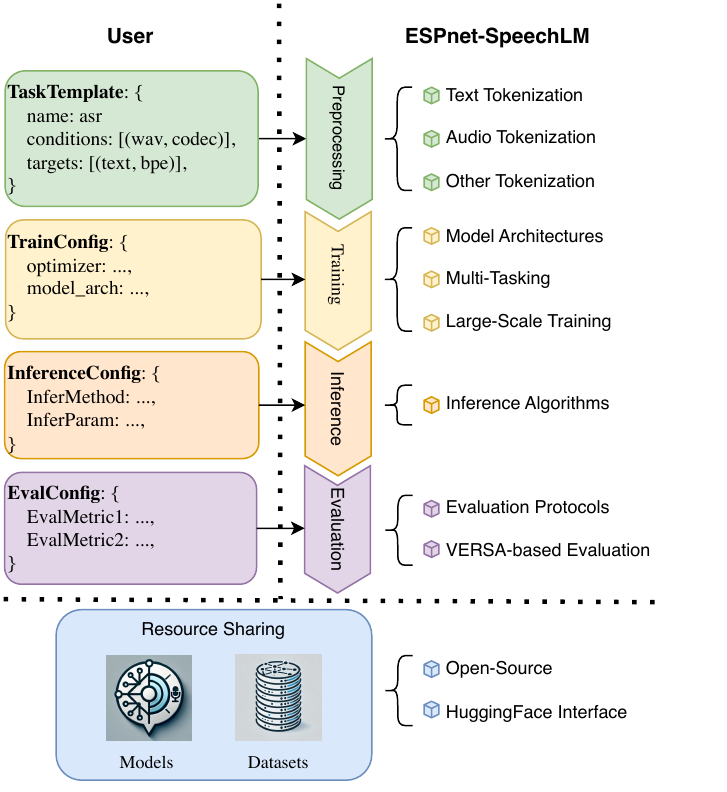}
    \caption{The overview of ESPnet-SpeechLM workflow. }
    \label{fig:overall_workflow}
    \vspace{-10pt}
\end{figure}

\begin{table*}[t]
    \centering
    \scalebox{0.65}{
    \begin{tabular}{l|c|c|c|c|c|c|c|c|c|c}
    \toprule
             &   \multicolumn{5}{|c|}{Open-Source Level} & \#Released & \multirow{2}{*}{\#Tasks} & \#Tokenizer & \#Tokenizer & \multirow{2}{*}{\#Architectures} \\
    Codebase &   Data & Train & Infer. & Eval. & Weights & Models     &   &   Types   & Choices    & \\
    \hline\hline
    VoxtLM      \cite{voxtlm}     & \checkmark & \checkmark & \checkmark & \checkmark & \checkmark & 1  & 4    & 2  & 2 & 1 \\
    UniAudio    \cite{uniaudio}   & \checkmark & \checkmark & \checkmark & \checkmark & \checkmark & 1  & 11   & 5  & 5 & 1 \\
    Moshi       \cite{moshi}      &            &            & \checkmark &            & \checkmark & 13 & N/A  & 2  & 2 & 1  \\
    Mini-Omni   \cite{mini_omni}  &            &            & \checkmark &            & \checkmark & 1  & N/A  & 2  & 2 & 1  \\
    GLM-4-Voice \cite{glmvoice}   &            &            & \checkmark &            & \checkmark & 3  & N/A  & 2  & 2 & 1  \\
    \hline\hline
    ESPnet-SpeechLM (this work)        & \checkmark & \checkmark & \checkmark & \checkmark & \checkmark & 3  & 15   & 10 & N/A & 4 \\
    \bottomrule
    \end{tabular}}
    \caption{Comparison between ESPnet-SpeechLM and other open-sourced SpeechLM codebases. For open-ended SpeechLM dialogue systems, the \#Tasks are not well-defined and left N/A. 
    ESPnet-SpeechLM provides multiple interfaces to bridge a massive number of tokenizer choices and the exact number is also left N/A. 
    Details of the supported features in ESPnet-SpeechLM are in Tab.\ref{tab:features}. 
    Information as of Dec 2024.
    }
    \label{tab:comparison}
    \vspace{-10pt}
\end{table*}

The advent of large language models (LLMs) has significantly advanced machine intelligence, particularly in the text domain \cite{gpt4, llama}. As research expands beyond text, LLMs are increasingly applied to multimodal scenarios \cite{mlm_survey, gpt4o, vita}, such as speech \cite{slm_survey1, slm_survey2} and vision \cite{vlm_survey}, with the aim of achieving higher-level intelligence and enhancing human-computer interactions. Within this context, Speech Language Models (SpeechLMs) have emerged as a powerful paradigm addressing challenges unique to speech processing.

SpeechLMs have demonstrated remarkable progress across a variety of speech tasks, including zero-shot generalization \cite{valle}, low-resource modeling \cite{speartts}, multi-task learning \cite{voxtlm, uniaudio}, instruction following \cite{instruct_following}, real-time interaction \cite{moshi, mini_omni}, and emergent abilities \cite{uniaudio1.5}. Similarly to text-based LLMs \cite{llm_scaling}, SpeechLMs benefit from scaling data volume, parameter size, and computational resources \cite{scaling}. These advances have fueled a growing interest in SpeechLM research within the speech and language processing community.

However, despite these advances, the development of SpeechLMs remains a complex and resource-intensive endeavor \cite{moshi}. Building such models requires significant expertise and effort across diverse tasks, from data preparation to training, inference, and evaluation. To address these challenges and democratize SpeechLM research, we introduce ESPnet-SpeechLM, an open-source toolkit designed to streamline and accelerate SpeechLM development.

ESPnet-SpeechLM unifies speech tasks under a sequential modeling framework and organizes the SpeechLM development process into a standardized workflow. As illustrated in Fig.\ref{fig:overall_workflow}, users begin by defining a custom task template, followed by configuring key parameters. The toolkit then automates all phases of the pipeline: preprocessing, training, inference, and evaluation (\S\ref{workflow}). This modular workflow supports a wide range of design choices, including tokenization methods, model architectures, dynamic multi-tasking, etc. In addition, ESPnet-SpeechLM provides a HuggingFace-compatible interface for sharing datasets and models (\S\ref{feature}). The toolkit is fully open-source, ensuring reproducibility and accessibility.

To showcase its versatility, we present several use cases demonstrating the scalability and efficiency of ESPnet-SpeechLM. These include building competitive SpeechLM-based automatic speech recognition (ASR) and text-to-speech (TTS) systems on datasets exceeding 200k hours of speech-text paired data (\S\ref{asrtts_exp}). We also detail the creation of a 1.7B-parameter multi-task SpeechLM, pre-trained on ASR, TTS, TextLM, and AudioLM tasks, by leveraging 240 billion \textit{text tokens or audio frames} (\S\ref{multitask_exp}).

\vspace{-3pt}
\section{Related Work} \label{related_work}
\vspace{-3pt}

The ESPnet-SpeechLM toolkit builds upon prior works in two main directions:

\noindent\textbf{Text LLM ecosystem:} 
Some popular development tools in text LLM ecosystems can be generalized to any large-scale sequential modeling task, which means they are also suitable for SpeechLM training. Examples of this include DeepSpeed \cite{deepspeed} and FlashAttention \cite{flashattention}.
To preserve text capability, it is common to initialize SpeechLMs from pre-trained text LLMs, which can rely on open-source platforms like HuggingFace Transformers\footnote{\url{https://github.com/huggingface/transformers}}. These tools are integrated into ESPnet-SpeechLM. 
We also noticed that current text LLM training frameworks \cite{megatron, llamafactory} provide limited support for speech features. Our toolkit is presented as a supplement in this direction.

\noindent\textbf{Open-Sourced SpeechLMs and Speech Toolkits:} Current research on SpeechLMs and their transparency has been significantly advanced by prior open-source SpeechLM research works \cite{glmvoice, mini_omni, moshi, uniaudio, voxtlm}.
SpeechLM research also greatly benefits from general speech processing and sequence-to-sequence modeling toolkits \cite{espnet, speechbrain, amphion, s3prl, nemo, fairseq}, as they provide a wide range of components applicable to SpeechLM development. 
ESPnet-SpeechLM is presented as a combination of cutting-edge SpeechLM research and well-established speech processing techniques within the open-sourced community. 
More specifically, it is built upon the existing ESPnet \cite{espnet} codebase to better exploit prior community efforts and compare with existing non-SpeechLM works.
We summarize ESPnet-SpeechLM and related codebases in Tab.\ref{tab:comparison}.

\section{ESPnet-SpeechLM Toolkit}
This section outlines the hierarchical design of the ESPnet-SpeechLM toolkit. We first introduce the fundamental concepts of SpeechLMs in \S\ref{background} followed by a detailed description of the ESPnet-SpeechLM workflow in \S\ref{workflow}. Lastly, we highlight key features of our toolkit in \S\ref{feature}.

\subsection{Speech Language Model} \label{background}
Speech tasks can be generically formulated as predicting target sequences $\mathbf{y} = [\mathbf{y}_1, ..., \mathbf{y}_N]$ given input conditions $\mathbf{x} = [\mathbf{x}_1, ..., \mathbf{x}_M]$, where each $\mathbf{x}_m$ and $\mathbf{y}_n$ represents individual data items. $M$ and $N$ stand for the number of data items in conditions and targets, respectively.
E.g., for ASR, $\mathbf{x}_1$ is the input speech; $\mathbf{y}_1$ is the corresponding transcription. 
Commonly, the training objective is to maximize the posterior $P(\mathbf{y}|\mathbf{x})$.

ESPnet-SpeechLM uniformly frames speech tasks as sequential modeling problems using auto-regressive prediction over discrete token sequences within a decoder-only Transformer \cite{transformer}.
Specifically, all data items $\mathbf{x}_m$ and $\mathbf{y}_n$ are first tokenized into discrete token sequences $\mathbf{x}_m^{\text{d}}$ and $\mathbf{y}_n^{\text{d}}$. 
Then, the spliced sequence $\mathbf{s}^{\text{d}} = [\mathbf{x}_1^{\text{d}}, ..., \mathbf{x}_M^{\text{d}}, \mathbf{y}_1^{\text{d}}, ..., \mathbf{y}_N^{\text{d}}]$ serves as the input for model training. 
Cross-entropy loss optimization over $\mathbf{y}_1^{\text{d}}, ..., \mathbf{y}_N^{\text{d}}$ approximates the objective $P(\mathbf{y}|\mathbf{x})$. 
Predicting $\hat{\mathbf{y}}_1^{\text{d}}, ..., \hat{\mathbf{y}}_N^{\text{d}}$ based on the conditions $\mathbf{x}_1^{\text{d}}, ..., \mathbf{x}_M^{\text{d}}$ and then detokenizing them into  $\hat{\mathbf{y}}_1, ..., \hat{\mathbf{y}}_N$ yield the final system prediction. 

ESPnet-SpeechLM specifically supports multi-stream language models, i.e., $\mathbf{s}^{\text{d}}\in\mathbb{N}^{T \times n_q}$ where $T$ stands for the sequence length and $n_q$ stands for the number of streams (See Fig.\ref{fig:template}). 
This capability is especially critical for audio codec models \cite{encodec, soundstream}, which encode each audio frame into multiple tokens. 
Padding tokens are added to align non-audio data when splicing $\mathbf{s}^{\text{d}} = [\mathbf{x}_1^{\text{d}}, ..., \mathbf{x}_M^{\text{d}}, \mathbf{y}_1^{\text{d}}, ..., \mathbf{y}_N^{\text{d}}]$. 
These multi-stream models require specialized design considerations discussed in \S\ref{feature}.

\subsection{ESPnet-SpeechLM Workflow} \label{workflow}
The ESPnet-SpeechLM workflow begins with single-task scenarios and extends naturally to multitask training. In the following, we introduce the concept of the task template (\S\ref{template}) and describe the end-to-end pipeline from preprocessing to evaluation (\S\ref{preprocessing}-\ref{evaluation}). Multitasking is described in \ref{multi-task}.

\subsubsection{Task Template} \label{template}
As in \S\ref{background}, regardless of the exact sequence $\mathbf{s}^{\text{d}}$, the SpeechLM performs sequential modeling indiscriminately. It is the definition of conditions $\mathbf{x}$, targets $\mathbf{y}$, and the corresponding tokenization methods that give the distinctive composition of $\mathbf{s}^{\text{d}}$ and thus the support of different tasks within SpeechLMs. 


\begin{figure}
    \centering
    \includegraphics[width=\linewidth]{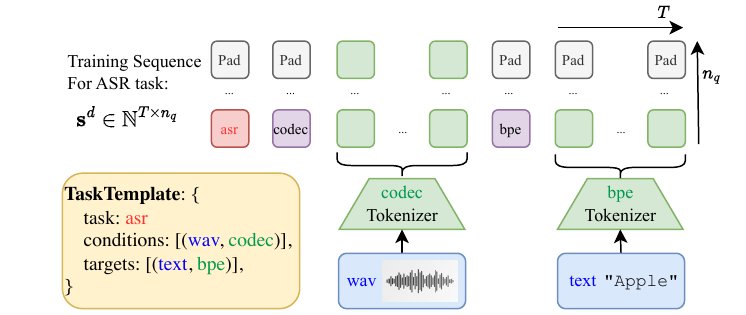}
    \caption{The training sequence $\mathbf{s}^{\text{d}}$ is assembled based on the \textit{task template}, e.g. single-task ASR as depicted here. The sequence is multi-stream with an extra $n_q$-axis because the codec tokenizes each frame into multiple tokens.}
    \label{fig:template}
    \vspace{-10pt}
\end{figure}

To handle different speech tasks uniformly, ESPnet-SpeechLM defines each task using a \textit{task template}, which specifies the composition of the training sequence $\mathbf{s}^{\text{d}}$.
As shown in Fig.\ref{fig:template}, the task template defines the name of the \textcolor{red}{task}, the conditions, and the targets. For each data item $\mathbf{x}_m$ or $\mathbf{y}_n$, the \textcolor{blue}{item\_name} and \textcolor{ForestGreen}{tokenizer} are specified. The training sequence starts from the \textcolor{red}{task} identifier, followed by the tokenized sequences from all the data items.
For each data item, the first token is a \textcolor{Mulberry}{tokenizer indicator}; the raw data is tokenized by the specified \textcolor{ForestGreen}{tokenizer}. For single-stream data items and special tokens, \textcolor{Gray}{padding tokens} are added. With this template, the training sequences can be assembled automatically from the given raw data.

\subsubsection{Preprocessing} \label{preprocessing}
Preprocessing primarily involves tokenization.
As some tokenizers are neural network-based and heavy, it is more efficient to conduct the tokenization offline. 
The tokenization is handled automatically by ESPnet-SpeechLM after receiving a folder for each train/valid/test set as follows. Specifically, an index file is provided for each data item, with the format of \texttt{example-id content} in each line. The name of these index files should correspond to the \textcolor{blue}{item\_name} in the task template. 


\begin{quote}
\small \textsf{
\textit{(folder)} train \\
     \textcolor{white}{--------}    |-- \textit{(file)} \textcolor{blue}{wav}   \\
     \textcolor{white}{------------}    |-- \textit{(line)} \texttt{example-id1 path-to-wav1} \\ 
     \textcolor{white}{------------}    |-- \textit{(line)} \texttt{example-id2 path-to-wav2} \\ 
     \textcolor{white}{--------}    |-- \textit{(file)} \textcolor{blue}{text}  \\
     \textcolor{white}{------------}    |-- \textit{(line)} \texttt{example-id1 text1} \\ 
     \textcolor{white}{------------}    |-- \textit{(line)} \texttt{example-id2 text2} \\ 
}
\vspace{-10pt}
\end{quote}

ESPnet-SpeechLM processes these files to generate a unified \texttt{data.json} file for each dataset, which contains tokenized results and metadata. \texttt{data.json} is the data format in both training and evaluation. 
During preprocessing, all tokenizers in use are detected, and a joint vocabulary is constructed automatically.

\begin{table*}[t]
    \centering
    \caption{Summary of supported features in ESPnet-SpeechLM toolkit}
    \scalebox{0.6}{
    \begin{tabular}{c|c|l|l}
    \toprule
    \multirow{3}{*}{Task Templates} &  \multicolumn{3}{|l}{TextLM, AudioLM, Text-to-Speech, Automatic Speech Recognition, Machine Translation,} \\
     & \multicolumn{3}{|l}{Speech-to-Text Translation, Speech-to-Speech Translation, Text-to-Audio, Text-to-Music, Audio Caption, } \\ 
     & \multicolumn{3}{|l}{Text-to-Music, Singing Voice Synthesis, Speech Enhancement, Target Speaker Extraction, Visual TTS} \\
    \midrule
    \multirow{11}{*}{Tokenization} & \multirow{3}{*}{Text} & Subword & \href{https://github.com/google/sentencepiece}{SentencePiece}, \href{https://github.com/huggingface/tokenizers}{HuggingFace Tokenizers} \\
    \cmidrule{3-4}
                                  &                       & G2P & \href{https://github.com/espnet/espnet/blob/master/espnet2/text/phoneme_tokenizer.py}{30 choices} \\ 
    \cmidrule{2-4}
                 & \multirow{6}{*}{Audio}  & \multirow{2}{*}{Codec}         & ESPnet-Codec \cite{espnet_codec}, DAC \cite{dac}, \\
                 &                         &                                & Encodec \cite{encodec}, UniAudio \cite{uniaudio} \\ 
                 \cmidrule{3-4}
                 &                         & \multirow{2}{*}{SSL}           & XEUS \cite{xeus}, S3PRL \cite{s3prl}, \\
                &                          &                                & FairSeq \cite{fairseq} \\
                 \cmidrule{3-4}
                 &                         & Codec\_SSL    & Combine Codec and SSL frame-by-frame \\
    \cmidrule{2-4}
                 & \multirow{2}{*}{Others} & \multicolumn{2}{|l}{Music Score \cite{muskits}, Vision Token \cite{avhubert}, } \\ 
                 &                         & \multicolumn{2}{|l}{Classification Label, Speaker Identity, LLM Embeddings} \\ 
    \midrule 
\multirow{5}{*}{Modeling \& Training}  &  Transformer Body & \multicolumn{2}{|l}{ESPnet Builtin, \href{https://huggingface.co/docs/transformers/en/model_doc/auto\#transformers.AutoModelForCausalLM}{HuggingFace \texttt{AutoModelForCausalLM}}} \\
    \cmidrule{2-4}
    & Multi-Stream   & \multicolumn{2}{|l}{Vall-E \cite{valle}, MultiScale-Transformer \cite{uniaudio}} \\
    & Language Model & \multicolumn{2}{|l}{Parallel Interleave, Delay Interleave \cite{musicgen}}\\
    \cmidrule{2-4}
    & Efficiency       & \multicolumn{2}{|l}{DeepSpeed \cite{deepspeed}, FlashAttention \cite{flashattention}, \href{https://github.com/linkedin/Liger-Kernel}{Liger-Kernel}} \\
    \midrule
     Inference        & \multicolumn{3}{|l}{Greedy Search, Beam Search, Top-k Sampling, Top-p Sampling} \\ 
    \midrule 
     Evaluation       & \multicolumn{3}{|l}{VERSA \cite{espnet_codec}, with 61 evaluation metrics for speech and audio} \\ 
     \midrule
     \multirow{2}{*}{Sharing} & Task Template & \multicolumn{2}{|l}{\href{https://github.com/espnet/espnet}{ESPnet GitHub Repository}} \\ 
     \cmidrule{2-4}
     & Datasets \& Models & \multicolumn{2}{|l}{\href{https://huggingface.co/espnet}{ESPnet HuggingFace Hub}} \\
     \bottomrule
    \end{tabular}}
    \label{tab:features}
    \vspace{-5pt}
\end{table*}

\subsubsection{Training} \label{training}
The training behavior of ESPnet-SpeechLM is specified by a configuration file. 
Besides common training configurations like optimization, batch size, and distributed training setup, the toolkit also supports flexible model architecture configurations for SpeechLM development.

ESPnet-SpeechLM provides multiple implementations of multi-stream language models \cite{valle, musicgen, uniaudio}. 
The implementations of multi-stream language models all rely on the Transformer body implementation. We provide the ESPnet built-in Transformer implementation to maximize flexibility;
alternatively, we support any \texttt{AutoModelForCausalLM} from HuggingFace Transformers to leverage pre-trained text LLMs
Also, following \citet{moshi}, custom weights can be provided during loss computing to balance the tokens from different tokenization methods. This is usually to guarantee one audio frame has the same loss weight as one non-audio token. 
Lastly, in addition to applying the cross-entropy loss, the toolkit also supports reinforcement learning from human feedback (RLHF) for SpeechLMs (see \cite{dpo} for details). 

\subsubsection{Inference} \label{inference}
For each of the supported multi-stream language models, we provide multiple inference methods, such as greedy search, beam search, and top-k/top-p sampling. Our implementation also allows multiple heuristics like the min/max generation length. One important heuristic is essential to SpeechLM: unlike text LLMs that only predict text, SpeechLMs need to know the modality of the current predicting target, so that tokens from other modalities can be filtered out to avoid invalid predictions. The current modality is known from the most recent tokenizer indicator (\S\ref{template}), and will switch when a new tokenizer indicator is predicted.



\subsubsection{Evaluation} \label{evaluation}
We create an evaluation script for each supported task. Within these scripts, we consistently adopt the VERSA \footnote{\url{https://github.com/shinjiwlab/versa}}, a comprehensive collection of >60 speech and audio evaluation metrics \cite{espnet_codec}. 
Besides the existing evaluation scripts, a model in a new task can be evaluated simply by specifying the metrics, the inference results, and the reference (if needed). 

\subsubsection{Multitasking} \label{multi-task}
To build SpeechLMs with versatile functionalities, ESPnet-SpeechLM flexibly supports multitasking. 
As the SpeechLMs have the same modeling procedure for all tasks, achieving multitasking training is to fuse the training sequences from different tasks in the mini-batches. 
Similar to single-task training, for each task, the task template definition (\S\ref{template}) and preprocessing (\S\ref{preprocessing}) are completed separately, which gives multiple tokenized datasets and the corresponding \texttt{data.json} files\footnote{
  Especially, these preprocessing works are easy to distribute and are suitable for collaborative works.
}. 
The data loader accepts a list of \texttt{data.json} and fuses these datasets before training, which allows the users to dynamically change the multitasking data setups. 
Mini-batches are sampled from the fused datasets during training. Additionally, the sampling ratio among these datasets is adjustable to emphasize some specific data portions.

\begin{table}[t]
    \centering
        \caption{English ASR performance (WER\%$\downarrow$) comparison among whisper \cite{whisper}, OWSM v3.1 \cite{owsm3.1} and ESPnet-SpeechLM ASR (ours). All results are derived from the greedy search.}
    \scalebox{0.56}{
    \begin{tabular}{l|c|c|c|c|c}
    \toprule
    &  \multicolumn{2}{c|}{Whisper} & \multicolumn{2}{c|}{OWSM v3.1} & \multirow{2}{*}{ours} \\ 
    \cmidrule{2-5}
                           & small & medium & small & medium &  \\ 
    \cmidrule{2-5}
    Test sets                  & 244M  & 769M   & 367M  & 1.02B  & 442M \\
    \midrule 
    LS-Clean \cite{librispeech}              &  3.3   & 2.8    & 2.5   & 2.4    & \bf{1.9}  \\
    LS-Other \cite{librispeech}             &  7.7   & 6.5    & 5.8   & 5.0    & \bf{4.6}  \\
    MLS  \cite{mls}                 &  9.1   & 10.2   & 8.1   & \bf{7.1}    & 7.2  \\
    TEDLIUM3 \cite{tedlium3}             &  \bf{4.6}   & 5.1    & 5.0   & 5.1    & 5.7  \\
    WSJ  \cite{wsj}                 &  4.3   & \bf{2.9}    & 3.8   & 3.5    & 5.2  \\ 
    FLEURS   \cite{FLEURS}             &  9.6   & \bf{6.4}    & 10.3  & 9.0    & 7.7  \\
    \midrule
    Avg.                   &  6.4   & 5.7    & 5.9   & \bf{5.4}    & \bf{5.4}  \\
    \bottomrule
    \end{tabular}}
    \label{tab:asr}
    \vspace{-8pt}
\end{table}

\subsection{Supported Features} \label{feature}
We summarize the core configurable features in ESPnet-SpeechLM workflow in Tab.\ref{tab:features} and highlight them as follows:

\noindent\textbf{Tokenization: } For text, we support subword models and grapheme-to-phoneme (G2P) tools, with an emphasis on HuggingFace tokenizers. For audio tokenization, we support both audio codec models and self-supervised learning (SSL) tokens. We provide multiple options for these two tokenization methods, with an emphasis on ESPnet-Codec \cite{espnet_codec} and XEUS \cite{xeus}. 
Additionally, we find that concatenating codec and SSL tokens frame-by-frame behaves well in both speech understanding and generation. 
Besides text and audio, these multi-modal models can leverage information from auxiliary modalities, such as music score \cite{muskits}, vision token \cite{avhubert}, classification labels (e.g., bool, time-stamp), speaker-identity and the continuous LLM embeddings. 

\noindent\textbf{Training: } As in \S\ref{training}, for the Transformer body, we provide the ESPnet built-in implementation as well as the HuggingFace Transformers implementation. Upon the Transfomer, we support 4 distinctive multi-stream language model implementations \cite{valle, uniaudio, musicgen}. For training efficiency, we leverage DeepSpeed \cite{deepspeed}, FlashAttention \cite{flashattention} and Liger-Kernel\footnote{\url{https://github.com/linkedin/Liger-Kernel}}. These modules enable us to achieve model FLOPs utility (MFU) \cite{mfu} as high as 35\% with multi-node training using NVIDIA H100 GPUs.

\noindent\textbf{Inference, Evaluation, and Sharing: } For all supported architecture, we provide all 4 inference methods. VERSA provides more than 60 speech-related evaluation metrics. To ensure transparency and reproducibility, the code and task templates are released through the ESPnet GitHub repository; tokenized datasets and pre-trained models are released through ESPnet Huggingface Hub.

\begin{table}[t]
    \centering
    \caption{
    TTS performance on full LibriSpeech Test-Clean \cite{librispeech}. 
    SPK\_SIM is measured only when zero-shot speaker prompting is supported. The speaker prompts are the same for all tests. All results from VERSA. No post-selection applied. $\star$ means third-party implementation.
    }
    \scalebox{0.6}{
    \begin{tabular}{l|c|c|c}
    \toprule
        Model & WER($\downarrow$) & SPK\_SIM($\uparrow$) & Proxy MOS($\uparrow$)  \\
        \midrule
        ChatTTS \cite{chattss2024}                     & 7.1 & - & 3.52\\
        CosyVoice \cite{du2024cosyvoice}               & 5.0 & 0.51 & \bf{4.15} \\
        Parler-TTS \cite{lacombe-etal-2024-parler-tts} & 4.7 & - & 3.83 \\
        WhisperSpeech \cite{whisperspeech2024}         & 13.5 & - & 4.06 \\
        VallE-X $\star$ \cite{zhang2023speak}                  & 27.3 & 0.35 & 3.38 \\ 
        VallE 2 $\star$ \cite{valle2}                          & 27.8 & 0.46 & 3.65 \\ 
        \midrule
        ESPnet-SpeechLM TTS (ours)                     & \bf{3.1} & \bf{0.55} & 4.03 \\  
        \bottomrule
    \end{tabular}}
    \label{tab:tts}
    \vspace{-8pt}
\end{table}

\section{User Cases}

\begin{table*}[t]
    \centering
    \caption{Evaluation on the multitask pre-trained SpeechLM using ESPnet-SpeechLM and its comparison with prior text LLM, SpeechLMs, and Multimodal LMs. The numbers of competitors are from their own report unless marked by $\star$. - means unreported numbers. 
    }
    \scalebox{0.58}{
    \begin{tabular}{l|c|c|ccc|cccc|c}
    \toprule
     Task    &  & \multicolumn{1}{|c}{ASR} & \multicolumn{3}{|c}{TTS} & \multicolumn{4}{|c}{TextLM} & \multicolumn{1}{|c}{AudioLM} \\
     \midrule
    Metric & Size & WER($\downarrow$) & WER($\downarrow$) & SPK\_SIM($\uparrow$) & Proxy MOS($\uparrow$) & MMLU($\uparrow$) & ARC-C($\uparrow$) & HS($\uparrow$) & OBQA($\uparrow$) & Perplexity($\downarrow$) \\
    \midrule
    LLaMA-3.2 \cite{llama} & 1B  & - & - & - & - & 32.2 & 32.8 & 41.2 & 29.2$^\star$ & -  \\
    VoxtLM \cite{voxtlm}   & 1.3B & 2.7 / 6.5 & - & - & - & - & - & - & - & 40.9 \\ 
    Moshi \cite{moshi}   & 7B &   5.7 / - \phantom{0}   & 4.7 & - & - & 49.8 & -  & - & - & -  \\ 
    MiniOmni \cite{mini_omni} & 0.5B  & 4.5 / 9.7  & - & - & - & - & - &  - & - & -  \\
    VITA \cite{vita}       & 8x7B & 8.1 / 18.4 & - & - & - & 71.0 & - & - & - &  -  \\ 
    GLM-4-Voice \cite{glmvoice} & 9B   & 2.8 / 7.7 & 5.6  & - & - & - & - &  - &  - & -  \\
    \midrule\midrule
    ESPnet-SpeechLM (ours) & 1.7B  & 2.8 / 5.9 & 6.0 & 0.701 & 3.99 & 30.5 & 41.3 & 50.4 & 31.4 & 16.4 \\
    \bottomrule
    \end{tabular}}
    \label{tab:lm}
    \vspace{-5pt}
\end{table*}

This section provides several user cases to demonstrate the performance of SpeechLMs built from ESPnet-SpeechLM. 
We first build single-task SpeechLM-Style ASR and TTS models in \S\ref{asrtts_exp}.
As a highlight of this demo, we present a 1.7B pre-trained SpeechLM that covers 4 tasks similar to \cite{voxtlm}: ASR, TTS, text auto-regressive prediction (TextLM), and speech auto-regressive prediction (AudioLM) (\S\ref{multitask_exp}). These models are released through ESPnet HuggingFace Hub\footnote{\url{https://huggingface.co/espnet}}. 

\subsection{Experimental Setups} \label{setup}
\textbf{Model and Tokenization: } We consistently leverage the pre-trained text LLM, SmolLM2 series\footnote{\url{https://huggingface.co/HuggingFaceTB}}, for SpeechLM initialization. We adopt the 360M and 1.7B versions for single-task and multi-task models, respectively. 
We adopt delay interleave \cite{musicgen} as the multi-stream language model architecture.
In terms of tokenization, we adopt the Codec\_SSL method
(\S\ref{feature}) for speech representation.
To preserve full transparency and self-consistency, ESPnet-Codec\footnote{\url{https://huggingface.co/ftshijt/espnet\_codec\_dac\_large\_v1.4\_360epoch}} 
and XEUS\footnote{\url{https://huggingface.co/espnet/xeus}; K-Means tokenizer trained on its last layer of representation using 5k clusters} are adopted for codec and SSL tokenizers respectively. 

\noindent\textbf{Data, Training, and Inference: } We collect open-sourced data for all experiments. Our data contains 200k hours of speech and 115B tokens of text, most in English. When expanding speech data into ASR, TTS, and AudioLM tasks, this is equivalent to 240B \textit{text tokens or audio frames}. Detailed data composition is in Appendix \ref{sec:appendix2}. 
We balance the weights for text, SSL, and codec tokens as 1: 0.5: 0.0625\footnote{Each audio frame is represented by 1 SSL token and 8 codec tokens. This ratio is to ensure (1) the text tokens have the same weight as the audio frames, and (2) SSL tokens have the same weight as 8 codec tokens combined.}.
The training used 8/24 H100 GPUs for single/multi-task training. We use batch size as large as around 2M \textit{frames or tokens} and a constant learning rate of 2e-4, with 10k warmup steps. We train the model for 2 data passes. We use greedy search for ASR and top-k sampling for TTS ($k=30, temperature=0.8$). 

\noindent\textbf{Evaluation: } Following \cite{voxtlm, dpo}, we test word error rate (WER) for ASR; ASR WER, Speaker Similarty and Proxy MOS for TTS; perplexity for AudioLM. We measure the TextLM ability using popular metrics like MMLU \cite{mmlu}, ARC-Challenge (ARC-C) \cite{arc}, HellaSwag (HS)\cite{hellaswag}, and OpenBookQA (OBQA) \cite{openbookqa}. 

\subsection{ASR and TTS Experiments} \label{asrtts_exp}
We evaluate our ASR system on multiple benchmarks and compare it with the popular open-sourced ASR models: whisper-v3-large \cite{whisper} and OWSM v3.1-medium \cite{owsm3.1}. As suggested in Tab.\ref{tab:asr}, our SpeechLM-based ASR system achieves comparable results in English with these two popular speech recognizers even using much fewer parameters. 
In Tab.\ref{tab:tts}, we compare the ESPnet-SpeechLM TTS system with other discrete-based TTS systems\footnote{For VallE-X and VallE 2, we use the third-party implementations: \url{https://huggingface.co/Plachta/VALL-E-X/resolve/main/vallex-checkpoint.pt}, \url{https://huggingface.co/amphion/valle}}. 
The results suggest our TTS system achieves decent performance on all evaluation metrics. 

\subsection{Multi-Task Experiments} \label{multitask_exp}
We demonstrate the performance of our multitask pre-trained SpeechLM in Tab.\ref{tab:lm}. 
Compared with other SpeechLMs \cite{voxtlm, mini_omni, glmvoice, llamaomni} and multimodal LMs \cite{vita}, our pre-trained model still preserves decent ASR, TTS and AudioLM performance even with limited parameter budget. 
In terms of text capability, the pre-trained model preserves close performance compared with the text-only LLM LLaMA-3.2-1B \cite{llama}. 

\section{Future Works}
We will continue the development of the ESPnet-SpeechLM toolkit, such as supporting more tokenization methods, more task templates, more modeling options, and LLM inference engines \cite{vllm}.
We are also interested in applying this toolkit to our SpeechLM research. 
For pre-training, we are interested in larger-scale models and models that can capture rich paralinguistic information in speech.
For post-training, we are interested in achieving conversational interactions, speech-based instruction following ability, and even agent-alike behaviors. 
Our plan also includes real-time and duplex design, HFRL for SpeechLM and SpeechLMs that trained from flat start.


\section{Conclusion}
This demo presents ESPnet-SpeechLM, a toolkit that covers the whole workflow of speech language model development, with comprehensive support in multiple design choices. 
We also provide user cases for both single-task and multi-task training, showing competitive performance with other models in the market.
The toolkit promises to keep full transparency in data, code, recipes, and pre-trained models.

\bibliography{custom}

\begin{thebibliography}{63}
\providecommand{\natexlab}[1]{#1}

\bibitem[{2Noise(2024)}]{chattss2024}
2Noise. 2024.
\newblock \href {https://github.com/2noise/ChatTTS} {Chattts: A generative speech model for daily dialogue.}
\newblock Available at \url{https://github.com/2noise/ChatTTS}.

\bibitem[{Achiam et~al.(2023)}]{gpt4}
Josh Achiam et~al. 2023.
\newblock Gpt-4 technical report.
\newblock \emph{arXiv preprint arXiv:2303.08774}.

\bibitem[{Chen et~al.(2024{\natexlab{a}})}]{valle2}
Sanyuan Chen et~al. 2024{\natexlab{a}}.
\newblock Vall-e 2: Neural codec language models are human parity zero-shot text to speech synthesizers.
\newblock \emph{arXiv preprint arXiv:2406.05370}.

\bibitem[{Chen et~al.(2024{\natexlab{b}})}]{xeus}
William Chen et~al. 2024{\natexlab{b}}.
\newblock Towards robust speech representation learning for thousands of languages.
\newblock \emph{arXiv preprint arXiv:2407.00837}.

\bibitem[{Chowdhery et~al.(2023)}]{mfu}
Aakanksha Chowdhery et~al. 2023.
\newblock Palm: Scaling language modeling with pathways.
\newblock \emph{Journal of Machine Learning Research}, 24(240):1--113.

\bibitem[{Clark et~al.(2018)}]{arc}
Peter Clark et~al. 2018.
\newblock Think you have solved question answering? try arc, the ai2 reasoning challenge.
\newblock \emph{arXiv preprint arXiv:1803.05457}.

\bibitem[{Collabora(2024)}]{whisperspeech2024}
Collabora. 2024.
\newblock \href {https://github.com/collabora/WhisperSpeech} {Whisperspeech: A speech processing toolkit}.
\newblock Available at \url{https://github.com/collabora/WhisperSpeech}.

\bibitem[{Conneau et~al.(2022)}]{FLEURS}
Alexis Conneau et~al. 2022.
\newblock {FLEURS: Few-Shot Learning Evaluation of Universal Representations of Speech}.
\newblock In \emph{SLT}.

\bibitem[{Copet et~al.(2024)}]{musicgen}
Jade Copet et~al. 2024.
\newblock Simple and controllable music generation.
\newblock \emph{Advances in Neural Information Processing Systems}, 36.

\bibitem[{Cuervo and Marxer(2024)}]{scaling}
Santiago Cuervo and Ricard Marxer. 2024.
\newblock Scaling properties of speech language models.
\newblock \emph{arXiv preprint arXiv:2404.00685}.

\bibitem[{Cui et~al.(2024)}]{slm_survey1}
Wenqian Cui et~al. 2024.
\newblock Recent advances in speech language models: A survey.
\newblock \emph{arXiv preprint arXiv:2410.03751}.

\bibitem[{Dao(2023)}]{flashattention}
Tri Dao. 2023.
\newblock Flashattention-2: Faster attention with better parallelism and work partitioning.
\newblock \emph{arXiv preprint arXiv:2307.08691}.

\bibitem[{D{\'e}fossez et~al.(2022)}]{encodec}
Alexandre D{\'e}fossez et~al. 2022.
\newblock High fidelity neural audio compression.
\newblock \emph{arXiv preprint arXiv:2210.13438}.

\bibitem[{D{\'e}fossez et~al.(2024)}]{moshi}
Alexandre D{\'e}fossez et~al. 2024.
\newblock Moshi: a speech-text foundation model for real-time dialogue.
\newblock \emph{arXiv preprint arXiv:2410.00037}.

\bibitem[{Du et~al.(2024)}]{du2024cosyvoice}
Zhihao Du et~al. 2024.
\newblock Cosyvoice: A scalable multilingual zero-shot text-to-speech synthesizer based on supervised semantic tokens.
\newblock \emph{arXiv preprint arXiv:2407.05407}.

\bibitem[{Dubey et~al.(2024)}]{llama}
Abhimanyu Dubey et~al. 2024.
\newblock The llama 3 herd of models.
\newblock \emph{arXiv preprint arXiv:2407.21783}.

\bibitem[{Fang et~al.(2024)Fang, Guo, Zhou, Ma, Zhang, and Feng}]{llamaomni}
Qingkai Fang, Shoutao Guo, Yan Zhou, Zhengrui Ma, Shaolei Zhang, and Yang Feng. 2024.
\newblock Llama-omni: Seamless speech interaction with large language models.
\newblock \emph{arXiv preprint arXiv:2409.06666}.

\bibitem[{Fu et~al.(2024)}]{vita}
Chaoyou Fu et~al. 2024.
\newblock Vita: Towards open-source interactive omni multimodal llm.
\newblock \emph{arXiv preprint arXiv:2408.05211}.

\bibitem[{He et~al.(2024)}]{he2024emilia}
Haorui He et~al. 2024.
\newblock Emilia: An extensive, multilingual, and diverse speech dataset for large-scale speech generation.
\newblock \emph{arXiv preprint arXiv:2407.05361}.

\bibitem[{Hendrycks et~al.(2020)}]{mmlu}
Dan Hendrycks et~al. 2020.
\newblock Measuring massive multitask language understanding.
\newblock \emph{arXiv preprint arXiv:2009.03300}.

\bibitem[{Hernandez et~al.(2018)}]{tedlium3}
Fran{\c{c}}ois Hernandez et~al. 2018.
\newblock {TED-LIUM} 3: Twice as much data and corpus repartition for experiments on speaker adaptation.
\newblock In \emph{Speech \& Computer}, pages 198--208.

\bibitem[{Huang et~al.(2024)Huang, Cheng, Liu, Hao, Song, Xu, Yang, Liu, Zhang, Chai et~al.}]{huang2024opencoder}
Siming Huang, Tianhao Cheng, Jason~Klein Liu, Jiaran Hao, Liuyihan Song, Yang Xu, J~Yang, JH~Liu, Chenchen Zhang, Linzheng Chai, et~al. 2024.
\newblock Opencoder: The open cookbook for top-tier code large language models.
\newblock \emph{arXiv preprint arXiv:2411.04905}.

\bibitem[{Hurst et~al.(2024)}]{gpt4o}
Aaron Hurst et~al. 2024.
\newblock Gpt-4o system card.
\newblock \emph{arXiv preprint arXiv:2410.21276}.

\bibitem[{Kaplan et~al.(2020)}]{llm_scaling}
Jared Kaplan et~al. 2020.
\newblock Scaling laws for neural language models.
\newblock \emph{arXiv preprint arXiv:2001.08361}.

\bibitem[{Kharitonov et~al.(2023)}]{speartts}
Eugene Kharitonov et~al. 2023.
\newblock Speak, read and prompt: High-fidelity text-to-speech with minimal supervision.
\newblock \emph{Transactions of the Association for Computational Linguistics}, 11:1703--1718.

\bibitem[{Kuchaiev et~al.(2019)}]{nemo}
Oleksii Kuchaiev et~al. 2019.
\newblock Nemo: a toolkit for building ai applications using neural modules.
\newblock \emph{arXiv preprint arXiv:1909.09577}.

\bibitem[{Kumar et~al.(2024)}]{dac}
Rithesh Kumar et~al. 2024.
\newblock High-fidelity audio compression with improved rvqgan.
\newblock \emph{Advances in Neural Information Processing Systems}, 36.

\bibitem[{Kwon et~al.(2023)}]{vllm}
Woosuk Kwon et~al. 2023.
\newblock Efficient memory management for large language model serving with pagedattention.
\newblock In \emph{Proceedings of the ACM SIGOPS 29th Symposium on Operating Systems Principles}.

\bibitem[{Lacombe et~al.(2024)}]{lacombe-etal-2024-parler-tts}
Yoach Lacombe et~al. 2024.
\newblock Parler-tts.

\bibitem[{Li et~al.(2023)}]{yodas}
Xinjian Li et~al. 2023.
\newblock Yodas: Youtube-oriented dataset for audio and speech.
\newblock In \emph{2023 IEEE Automatic Speech Recognition and Understanding Workshop (ASRU)}, pages 1--8.

\bibitem[{Lu et~al.(2024)}]{instruct_following}
Ke-Han Lu et~al. 2024.
\newblock Developing instruction-following speech language model without speech instruction-tuning data.
\newblock \emph{arXiv preprint arXiv:2409.20007}.

\bibitem[{Maiti et~al.(2024)}]{voxtlm}
Soumi Maiti et~al. 2024.
\newblock Voxtlm: Unified decoder-only models for consolidating speech recognition, synthesis and speech, text continuation tasks.
\newblock In \emph{ICASSP 2024-2024 IEEE International Conference on Acoustics, Speech and Signal Processing (ICASSP)}, pages 13326--13330. IEEE.

\bibitem[{Mihaylov et~al.(2018)}]{openbookqa}
Todor Mihaylov et~al. 2018.
\newblock Can a suit of armor conduct electricity? a new dataset for open book question answering.
\newblock \emph{arXiv preprint arXiv:1809.02789}.

\bibitem[{Ott et~al.(2019)}]{fairseq}
Myle Ott et~al. 2019.
\newblock \href {https://doi.org/10.18653/v1/N19-4009} {fairseq: A fast, extensible toolkit for sequence modeling}.
\newblock In \emph{Proceedings of the 2019 Conference of the North {A}merican Chapter of the Association for Computational Linguistics (Demonstrations)}, pages 48--53, Minneapolis, Minnesota. Association for Computational Linguistics.

\bibitem[{Panayotov et~al.(2015)}]{librispeech}
Vassil Panayotov et~al. 2015.
\newblock Librispeech: an {ASR} corpus based on public domain audio books.
\newblock In \emph{2015 IEEE international conference on acoustics, speech and signal processing (ICASSP)}, pages 5206--5210. IEEE.

\bibitem[{Paul and Baker(1992)}]{wsj}
Douglas~B Paul and Janet Baker. 1992.
\newblock {The design for the Wall Street Journal-based CSR corpus}.
\newblock In \emph{Proc. Workshop on Speech and Natural Language}.

\bibitem[{Peng et~al.(2024{\natexlab{a}})}]{slm_survey2}
Jing Peng et~al. 2024{\natexlab{a}}.
\newblock A survey on speech large language models.
\newblock \emph{arXiv preprint arXiv:2410.18908}.

\bibitem[{Peng et~al.(2024{\natexlab{b}})}]{owsm3.1}
Yifan Peng et~al. 2024{\natexlab{b}}.
\newblock Owsm v3.1: Better and faster open whisper-style speech models based on e-branchformer.
\newblock In \emph{Interspeech 2024}, pages 352--356.

\bibitem[{Pratap et~al.(2020)}]{mls}
Vineel Pratap et~al. 2020.
\newblock {MLS}: A large-scale multilingual dataset for speech research.
\newblock In \emph{Interspeech}.

\bibitem[{Radford et~al.(2023)}]{whisper}
Alec Radford et~al. 2023.
\newblock Robust speech recognition via large-scale weak supervision.
\newblock In \emph{International conference on machine learning}, pages 28492--28518. PMLR.

\bibitem[{Rajbhandari et~al.(2020)}]{deepspeed}
Samyam Rajbhandari et~al. 2020.
\newblock Zero: Memory optimizations toward training trillion parameter models.
\newblock In \emph{SC20: International Conference for High Performance Computing, Networking, Storage and Analysis}, pages 1--16. IEEE.

\bibitem[{Ravanelli et~al.(2021)}]{speechbrain}
Mirco Ravanelli et~al. 2021.
\newblock \href {https://arxiv.org/abs/2106.04624} {{SpeechBrain}: A general-purpose speech toolkit}.
\newblock \emph{Preprint}, arXiv:2106.04624.
\newblock ArXiv:2106.04624.

\bibitem[{Shi et~al.(2022)}]{avhubert}
Bowen Shi et~al. 2022.
\newblock Learning audio-visual speech representation by masked multimodal cluster prediction.
\newblock In \emph{International Conference on Learning Representations}.

\bibitem[{Shi et~al.(2024)}]{espnet_codec}
Jiatong Shi et~al. 2024.
\newblock Espnet-codec: Comprehensive training and evaluation of neural codecs for audio, music, and speech.
\newblock \emph{arXiv preprint arXiv:2409.15897}.

\bibitem[{Shoeybi et~al.(2019)}]{megatron}
Mohammad Shoeybi et~al. 2019.
\newblock Megatron-lm: Training multi-billion parameter language models using model parallelism.
\newblock \emph{arXiv preprint arXiv:1909.08053}.

\bibitem[{Tian et~al.(2024{\natexlab{a}})}]{owsm3.2}
Jinchuan Tian et~al. 2024{\natexlab{a}}.
\newblock On the effects of heterogeneous data sources on speech-to-text foundation models.
\newblock In \emph{Interspeech 2024}, pages 3959--3963.

\bibitem[{Tian et~al.(2024{\natexlab{b}})}]{dpo}
Jinchuan Tian et~al. 2024{\natexlab{b}}.
\newblock Preference alignment improves language model-based tts.
\newblock \emph{arXiv preprint arXiv:2409.12403}.

\bibitem[{Vaswani et~al.(2017)}]{transformer}
Ashish Vaswani et~al. 2017.
\newblock Attention is all you need.
\newblock \emph{Advances in Neural Information Processing Systems}.

\bibitem[{Wang et~al.(2023)}]{valle}
Chengyi Wang et~al. 2023.
\newblock Neural codec language models are zero-shot text to speech synthesizers.
\newblock \emph{arXiv preprint arXiv:2301.02111}.

\bibitem[{Watanabe et~al.(2018)}]{espnet}
Shinji Watanabe et~al. 2018.
\newblock \href {https://doi.org/10.21437/Interspeech.2018-1456} {{ESPnet}: End-to-end speech processing toolkit}.
\newblock In \emph{Proceedings of Interspeech}, pages 2207--2211.

\bibitem[{Wu et~al.(2024)}]{muskits}
Yuning Wu et~al. 2024.
\newblock Muskits-espnet: A comprehensive toolkit for singing voice synthesis in new paradigm.
\newblock In \emph{Proceedings of the 32nd ACM International Conference on Multimedia}, pages 11279--11281.

\bibitem[{Xie and Wu(2024)}]{mini_omni}
Zhifei Xie and Changqiao Wu. 2024.
\newblock Mini-omni: Language models can hear, talk while thinking in streaming.
\newblock \emph{arXiv preprint arXiv:2408.16725}.

\bibitem[{Yang et~al.(2024{\natexlab{a}})}]{uniaudio1.5}
Dongchao Yang et~al. 2024{\natexlab{a}}.
\newblock Uniaudio 1.5: Large language model-driven audio codec is a few-shot audio task learner.
\newblock In \emph{The Thirty-eighth Annual Conference on Neural Information Processing Systems}.

\bibitem[{Yang et~al.(2024{\natexlab{b}})}]{uniaudio}
Dongchao Yang et~al. 2024{\natexlab{b}}.
\newblock Uniaudio: Towards universal audio generation with large language models.
\newblock In \emph{Forty-first International Conference on Machine Learning}.

\bibitem[{Yang et~al.(2021)}]{s3prl}
Shu-Wen Yang et~al. 2021.
\newblock {SUPERB: Speech Processing Universal PERformance Benchmark}.
\newblock In \emph{Proc. Interspeech 2021}, pages 1194--1198.

\bibitem[{Yin et~al.(2024)}]{mlm_survey}
Shukang Yin et~al. 2024.
\newblock A survey on multimodal large language models.
\newblock \emph{National Science Review}, page nwae403.

\bibitem[{Zeghidour et~al.(2021)}]{soundstream}
Neil Zeghidour et~al. 2021.
\newblock Soundstream: An end-to-end neural audio codec.
\newblock \emph{IEEE/ACM Transactions on Audio, Speech, and Language Processing}, 30:495--507.

\bibitem[{Zellers et~al.(2019)}]{hellaswag}
Rowan Zellers et~al. 2019.
\newblock Hellaswag: Can a machine really finish your sentence?
\newblock \emph{arXiv preprint arXiv:1905.07830}.

\bibitem[{Zeng et~al.(2024)}]{glmvoice}
Aohan Zeng et~al. 2024.
\newblock \href {https://arxiv.org/abs/2412.02612} {Glm-4-voice: Towards intelligent and human-like end-to-end spoken chatbot}.
\newblock \emph{Preprint}, arXiv:2412.02612.

\bibitem[{Zhang et~al.(2024{\natexlab{a}})}]{vlm_survey}
Jingyi Zhang et~al. 2024{\natexlab{a}}.
\newblock Vision-language models for vision tasks: A survey.
\newblock \emph{IEEE Transactions on Pattern Analysis and Machine Intelligence}.

\bibitem[{Zhang et~al.(2024{\natexlab{b}})}]{amphion}
Xueyao Zhang et~al. 2024{\natexlab{b}}.
\newblock Amphion: An open-source audio, music and speech generation toolkit.
\newblock In \emph{{IEEE} Spoken Language Technology Workshop, {SLT} 2024}.

\bibitem[{Zhang et~al.(2023)}]{zhang2023speak}
Ziqiang Zhang et~al. 2023.
\newblock Speak foreign languages with your own voice: Cross-lingual neural codec language modeling.
\newblock \emph{arXiv preprint arXiv:2303.03926}.

\bibitem[{Zheng et~al.(2024)}]{llamafactory}
Yaowei Zheng et~al. 2024.
\newblock Llamafactory: Unified efficient fine-tuning of 100+ language models.
\newblock In \emph{Proceedings of the 62nd Annual Meeting of the Association for Computational Linguistics (Volume 3: System Demonstrations)}, Bangkok, Thailand. Association for Computational Linguistics.

\end{thebibliography}

\appendix

\section{Data Details} \label{sec:appendix2}
\raggedbottom
The statistics of our training data are in Tab.\ref{tab:data}. We highlight as follows.

\noindent\textbf{Speech Data: } We collected 213k hours of open-source speech data and applied the following preprocessing.
(1) We only use the English subset of Emilia \cite{he2024emilia}; 
(2) We use the Emilia pipeline \cite{he2024emilia} to process the raw audio files in the English subset of Yodas \cite{yodas};
and (3) We only use the English subset of the OWSM \cite{owsm3.2} dataset. We exclude the MLS to avoid duplication. This data is not applied to TTS as the speaker identity is absent. 

\noindent\textbf{Text-Only Data: } The text pretraining dataset is a diverse and extensive collection of text data sourced from three primary domains, encompassing a total of 115.69 billion tokens. The largest segment, contributing 82.36 billion tokens, is derived from general web content (FineWeb-EDU\footnote{https://huggingface.co/datasets/HuggingFaceFW/fineweb-edu
}), offering a rich variety of information spanning numerous topics and styles, suitable for broad language understanding tasks. Complementing this is 20.56 billion tokens of multilingual text from the Multilingual CC News dataset\footnote{https://huggingface.co/datasets/intfloat/multilingual\_cc\_news}, which enhances the model's ability to comprehend and generate text across multiple languages, catering to global linguistic diversity. Lastly, 12.77 billion tokens are sourced from the OpenCoder Annealing Corpus~\cite{huang2024opencoder}, a code-centric dataset, which bolsters the model's proficiency in understanding and generating programming languages and technical instructions. Together, these datasets provide a balanced blend of general, multilingual, and technical data, creating a robust foundation for versatile language model capabilities.

\begin{table}[t]
    \centering
    \caption{Detailed composition of the training data used in this work}
    \scalebox{0.55}{
    \begin{tabular}{l|c|c|c|c|c}
    \toprule
        \multirow{2}{*}{Dataset}     & \multirow{2}{*}{\#Hours} & \multicolumn{4}{c}{Text Tokens or Audio Frames (B)} \\
        \cmidrule{3-6}
             &  & ASR & TTS & TextLM & AudioLM \\ 
        \midrule  
        LibriSpeech \cite{librispeech}       & 960 & 0.18 & 0.33  & - & 0.17 \\ 
        MLS         \cite{mls}               & 55k & 9.88 & 16.13 & - & 9.15 \\ 
        Emilia      \cite{he2024emilia}      & 50k & 9.10 & 18.16 & - & 8.33 \\
        Yodas       \cite{yodas}             & 80k & 12.14 & 23.75 & - &11.14 \\
        OWSM        \cite{owsm3.2}           & 27k & 5.29  & -     & - & 4.95 \\ 
        \midrule
        General Text                         &     &       &  & 82.36 &    \\ 
        Multilingual Text                    &     &       &  & 20.56 &    \\
        Code                                 &     &       &  & 12.77 &    \\ 
        \midrule
        Total                                & 213k & 36.59 & 58.37 & 115.69 & 33.74 \\ 
        \bottomrule
    \end{tabular}}
    \label{tab:data}
\end{table}

\pagebreak
\end{document}